\documentclass[10pt,twocolumn,letterpaper]{article}

\usepackage[pagenumbers]{cvpr} 

\newcommand{\prisma}{\includegraphics[width=1.4em]{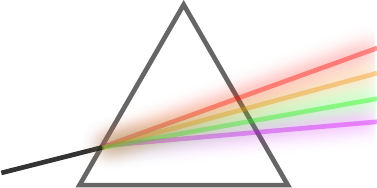}\textbf{Prisma} }

\usepackage[acronym]{glossaries}
\newacronym{ai}{AI}{artificial intelligence}
\newacronym{cnn}{CNN}{convolutional neural network}
\newacronym{fcnn}{FCNN}{fully-connected neural network}
\newacronym{ffnn}{FFNN}{feed-forward neural network}
\newacronym{gru}{GRU}{Gated Recurrent Unit}
\newacronym{lp}{LP}{linear programming}
\newacronym{milp}{MILP}{mixed integer linear programming}
\newacronym{mpt}{MPT}{Modern Portfolio Theory}
\newacronym{nlp}{NLP}{Natural Language Processing}
\newacronym{nn}{NN}{neural network}
\newacronym{ood}{OOD}{out-of-distribution}
\newacronym{pgd}{PGD}{projected gradient descent}
\newacronym{rnn}{RNN}{recurrent neural network}
\newacronym{sae}{SAE}{Sparse Autoencoder}
\newacronym{soa}{SoA}{state of the art}
\newacronym{tcn}{TCN}{temporal convolutional network}
\newacronym{vit}{ViT}{Vision Transformer}

\usepackage{fancyhdr}
\definecolor{cvprblue}{rgb}{0.21,0.49,0.74}
\usepackage[pagebackref,breaklinks,colorlinks,allcolors=cvprblue]{hyperref}
\usepackage{float}
\usepackage{svg}
\usepackage{easy-todo}
\def\paperID{10} 
\def\confName{CVPR}
\def\confYear{2025}

\title{\prisma: An Open Source Toolkit for Mechanistic Interpretability in Vision and Video}

\author{
Sonia Joseph$^{1,2, 3}$ \quad
Praneet Suresh$^{1,2}$ \quad
Lorenz Hufe$^{6}$ \\
Edward Stevinson$^{5}$ \quad
Robert Graham$^{2}$ \quad
Yash Vadi$^{1,4}$ \\
Danilo Bzdok$^{1,2}$ \quad
Sebastian Lapuschkin$^{6,7}$ \quad
Lee Sharkey$^{8}$ \\
Blake Aaron Richards$^{1,2}$ \\[0.3cm]
$^{1}$Mila Quebec \quad
$^{2}$McGill University \quad
$^{3}$Meta \quad
$^{4}$Université de Montréal \quad
$^{5}$Imperial College London \\
$^{6}$Fraunhofer Heinrich Hertz Institute \quad
$^{7}$Technological University Dublin \quad
$^{8}$Apollo Research\\[0.3cm]
\texttt{sonia.joseph@mila.quebec} \\ \small (Corresponding author)
}

\begin{document}
\maketitle

\begin{abstract}

Robust tooling and publicly available pre-trained models have helped drive recent advances in mechanistic interpretability for language models. However, similar progress in vision mechanistic interpretability has been hindered by the lack of accessible frameworks and pre-trained weights. We present Prisma\footnote{Access the codebase here: \url{https://github.com/Prisma-Multimodal/ViT-Prisma}}, an open-source framework designed to accelerate vision mechanistic interpretability research, providing a unified toolkit for accessing 75+ vision and video transformers; support for sparse autoencoder (SAE), transcoder, and crosscoder training; a suite of 80+ pre-trained SAE weights; activation caching, circuit analysis tools, and visualization tools; and educational resources. Our analysis reveals surprising findings, including that effective vision SAEs can exhibit substantially lower sparsity patterns than language SAEs, and that in some instances, SAE reconstructions can decrease model loss. Prisma enables new research directions for understanding vision model internals while lowering barriers to entry in this emerging field. 

\end{abstract}    

\section{Introduction}

Mechanistic interpretability has emerged as a powerful approach for understanding language models~\cite{bricken2023monosemanticity, kramar2024atpefficientscalablemethod}. This progress is driven by robust tools for circuit analysis and \gls{sae} training, enabled by features such as activation caching~\cite{nanda2022transformerlens, fiottokaufman2024nnsightndifdemocratizingaccess}, and the public release of pre-trained \glspl{sae} and training frameworks~\cite{bloom2024saetrainingcodebase, eleutherai2024sparsify}. 

In contrast, vision mechanistic interpretability has lagged behind partially due to limited access to tools, pre-trained weights, and standardized implementation code necessary for scalable research. 

\prisma bridges this gap by providing:

\begin{figure}
    \centering
    \includegraphics[width=\linewidth]{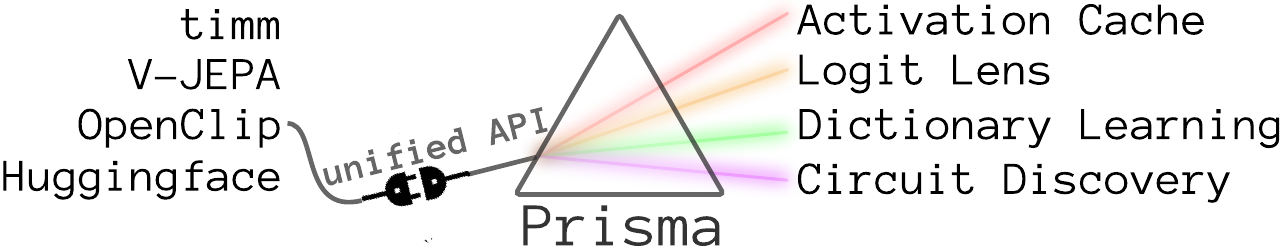}
    \caption{Prisma adapts popular vision model repositories for key mechanistic interpretability techniques.}
    \label{fig:enter-label}
\end{figure}

\begin{enumerate}
\item \textbf{Hooked Vision Transformers.} A unified interface for accessing and modifying \gls{vit} \cite{dosovitskiy2020image} and Video Transformer activations and weights across 75+ models from timm, OpenCLIP, and Huggingface, enabling diverse downstream interpretability tasks, similar to TransformerLens~\cite{nanda2022transformerlens}.
\item \textbf{SAEs, Transcoders, and Crosscoders.} Support for training and evaluating SAEs, and their variants, transcoders and crosscoders~\cite{cunningham2023sparse,bricken2023monosemanticity,dunefsky2025transcoders,lindsey2024sparse}. We also open source 80+ pre-trained SAE weights for CLIP-B and DINO-B \glspl{vit}~\cite{radford2021learning,ilharco_gabriel_2021_5143773,caron2021emerging,oquab2023dinov2} for all layers, transcoders for CLIP for all layers, with accompanying evaluations for all models (\cref{section: sae_evaluation_details}).
\item \textbf{Interpretability Tools.} A suite of tools, including methods for circuit analysis, attention head visualization, and logit lens techniques~\cite{nostalgebraist2020logitlens}.
\item \textbf{Educational Resources.} Tutorial notebooks, and documentation to ensure rapid onboarding. Additionally, toy vision transformers checkpoints are provided in the spirit of~\cite{elhage2022toy} to enable interpretability work in compute constrained environments.
\end{enumerate}

In this paper, we detail the contributions listed above. Then, using the Prisma toolkit, we explore some surprising findings about vision SAEs, including the high density of vision SAEs compared to language SAEs, and cases where injecting SAE activations into the forward pass can \textit{decrease} the original model's loss. We intend for the Prisma library to facilitate deeper investigations into vision transformer internals, accelerating the field of vision mechanistic interpretability.

\label{sec:intro}

\section{Background and Related Work}

\subsection{Mechanistic interpretability for vision and video}
Mechanistic interpretability in language has been flourishing, emphasizing causal circuit-tracing with sparse model representations to reverse engineer models \citep{elhage2022toy, wang2022interpretability, conmy2023towards, marks2024sparsefeaturecircuitsdiscovering}. Prior work on vision model interpretability has generally used different techniques from those that characterize current language mechanistic interpretability, such as feature visualization and attention analysis \citep{caron2021emerging, raghu2021vision, naseer2021intriguing}. Recently, mechanistic interpretability for vision models has been gaining traction \citep{joseph2024laying, gandelsman2024interpretingsecondordereffectsneurons, gandelsmanclipdecomposition, jucys2024interpretability,komorowski2023towards,achtibat2023attribution}, but research has been limited by the lack of scalable tooling, a gap which Prisma seeks to address.

Mechanistic interpretability for video remains limited, hindered by high compute costs and a lack of dedicated tools \cite{jucys2024interpretability, joseph2023mining}. Some methods automatically discover spatiotemporal concepts \cite{Kowal_2024_CVPR}, while others rely on proxy tasks to assess the model's use of dynamic information \cite{DBLP:journals/corr/abs-1807-06980, ilic2022appearancefreeactionrecognition}, sensitivity to scene bias \cite{choi2019cantidancemall, li2019repairremovingrepresentationbias, Li_2018_ECCV}, and difference between static and dynamic signals in intermediate representations \cite{kowal2024quantifyinglearningstaticvs}. We integrate two widely used video encoders, ViViT \cite{vivit2021} and V-JEPA \cite{bardes2024revisitingfeaturepredictionlearning}, into Prisma to facilitate interpretability research in video models.

\subsection{Mechanistic interpretability methods}

Neural networks represent more features than available neurons, making individual neurons polysemantic and difficult to interpret \citep{elhage2022toymodelssuperposition}. SAEs decompose single-layer representations into interpretable features \citep{cunningham2023sparse, bricken2023monosemanticity}, while transcoders extend this across layers for a more faithful approximation of feedforward computation \citep{templeton2024predicting, marks2024dictionary, dunefsky2024transcoders}. Crosscoders further track persistent features across layers and models \citep{lindsey2024sparse}.

Circuit analysis identifies interactions between model components to explain behaviors \cite{wang2022interpretability}. Techniques include activation patching \cite{heimersheim2024useinterpretactivationpatching}, which modifies activations in contrast pairs to isolate key components, and attribution patching \cite{nanda2023attributionpatching}, a faster gradient-based alternative. Interpretability has shifted from manual unit analysis to automated circuit discovery, with algorithms like ACDC \cite{conmy2023towards} for language and vision transformers \cite{rajaram2024automaticdiscoveryvisualcircuits}.

\subsection{Vision and video architectures}
Vision transformers (ViTs) process images by splitting them into tokenized patches, enabling self-attention analysis while achieving CNN-level performance \cite{dosovitskiy2020image}. CLIP \cite{radford2021learning} aligns visual and language representations via contrastive learning for zero-shot transfer, while DINO \cite{caron2021emerging} employs self-distillation for self-supervised learning.

Video transformers extend these concepts to sequential frames, including supervised models like ViViT \cite{vivit2021} and unsupervised models like V-JEPA \cite{bardes2024revisitingfeaturepredictionlearning}. We integrate these architectures into HookedViT, our vision model class that captures layer-wise activations for interpretability and sparse coder training.

\subsection{Open source mechanistic interpretability tools}

Mechanistic interpretability in language has a growing ecosystem of tools. TransformerLens enables circuit-style analysis of language models \cite{nanda2022transformerlens}, while NNsight supports model intervention but is not vision-specific \cite{fiottokaufman2024nnsightndifdemocratizingaccess}. CircuitsVis enables mechanistic interpretability visualizations \cite{cooney2023circuitsvis}. SAELens aids in training and analyzing \glspl{sae} \cite{bloom2024saetrainingcodebase}, and Gao et al. \cite{gao2024scalingevaluatingsparseautoencoders} released Top-K SAE training code, later extended by Eleuther \cite{eleutherai2024sparsify}. Gemma Scope provides open source SAEs on Gemma 2 \cite{lieberum2024gemmascopeopensparse}. For vision, Joseph and Nanda \cite{joseph2024laying} introduced a circuit-based framework for transformers, which Prisma extends with sparse coder training. However, no other open-source toolkits exist specifically for vision mechanistic interpretability, which differs from language mechanistic interpretability in tokenization, architecture, and analysis.


\subsection{Differences between vision and language mechanistic interpretability}

Vision and language transformers, despite sharing decoder-only architecture, differ fundamentally in ways that impact interpretability methods \cite{joseph2024laying, joseph2024vitprismanotebook}. Language input is discrete while vision input is continuous, creating differences in tokenization, processing, and decoding. ViTs use spatial patch tokenization without a canonical dictionary and use a learnable CLS token, which leads to distinct statistical properties between the two token types. Vision models typically employ contrastive learning or unsupervised approaches rather than next-token prediction. Due to the richness of visual representation, the outputs typically call for specialized decoding like cosine similarity or diffusion decoding instead of analyzing the next-token logit prediction. These architectural and functional differences call for specialized vision interpretability tools rather than adapting existing language interpretability libraries.

\section{The Prisma Open Source Library}

In this section, we introduce Prisma, an open-source toolkit designed to advance mechanistic interpretability in vision models. Building on the successes of interpretability tools in the language domain, Prisma adapts and extends these techniques to tackle the unique challenges encountered in vision. By retaining the naming conventions and hook interfaces established by TransformerLens \cite{nanda2022transformerlens}, Prisma enables researchers to easily transition between language and vision tooling.

\subsection{Prisma Codebase}
\subsubsection{Models Supported}

Our model registry integrates 75+ pretrained model configurations into HookedViT with TransformerLens-inspired activation caching for easy circuit analysis and SAE training. Support extends Hugging Face, OpenCLIP, and timm models (ViT, CLIP-VIT, DINO), and includes video-specific features like 3D tubelet embeddings for video models (ViViT, V-JEPA). We also adapt the Kandinsky ViT encoder with its pretrained SAE for steering diffusion models for image generation. The model registry structure enables convenient adoption of freshly released open-source models from the broader open-source community.

\subsubsection{Codebase Structure}
Built around HookedViT, key components include:

\begin{itemize}
    \item \textbf{Circuit Analysis:} Supports ablations via forward pass interventions. The HookedSAEViT module enables sparse feature circuit analysis through layer replacement with multiple SAEs or sparse coders. The HookedViT module integrates with ACDC for automatic circuit discovery.

    \item \textbf{Sparse Coder Architectures:} Enables training SAEs, transcoders, and crosscoders from HookedViT representations, with ReLU~\cite{cunningham2023sparse,bricken2023monosemanticity}, Top-K~\cite{gao2024scaling}, and JumpReLU~\cite{rajamanoharan2024jumpingaheadimprovingreconstruction}, and Gated~\cite{rajamanoharan2024improvingdictionarylearninggated} implementations.

    \item \textbf{Configurable SAE Training:} Parameters for encoder and decoder initialization~\cite{DictionaryOptimisation2023Conerly}, activation normalization~\cite{ImprovementDictionary2023Templeton}, ghost gradients~\cite{GhostGrads2023JermynTempleton}, dead feature resampling~\cite{bricken2023monosemanticity}, and early stopping criteria. We have options for activation caching for latency improvement~\cite{OpenSourceDictionaries2023Marks, bloom2024saetrainingcodebase}, or alternatively, on-the-fly computation for VRAM optimization~\cite{eleutherSparsify}. We chose to make the config expressive rather than minimal to accommodate emerging vision SAE best practices.
    
    \item \textbf{SAE Evaluation Tools:} Measures metrics like sparsity (L0), reconstruction and substitution losses~\cite{bricken2023monosemanticity}, and identifies maximally activating images~\cite{lim2024sparse} for trained sparse coders.
    
    \item \textbf{Visualization:} Provides logit lens and attention visualization for intermediate activations.
\end{itemize}


\section{Prisma Open Source Sparse Coders}
\label{section:saetrainingddetails}

We release \glspl{sae} for all layers of the CLIP and DINO base models and transcoders for CLIP base trained on ImageNet1k (~\cref{tab:models_structured}). For CLIP, we provide three token configurations: all patches, CLS-only, and spatial tokens, with standard ReLu and Top-K ($k = 64$) variants.
Key training config details include:

\begin{itemize}
    \item Expansion Factor: x64, mapping a 768-dimensional activation space to a dictionary of 49,152 features.
    \item Weight Initialization: Encoder weights are set as the transpose of the decoder weights.
    \item Training: Conducted on ImageNet1k \cite{deng2009imagenet} for one epoch using the Adam optimizer \cite{kingma2014adam}. Learning rates were swept from $1\times10^{-5}$ to $1\times10^{-1}$ with a cosine annealing schedule preceeded by a 200-step warmup and a batch size of 4096.
    \item L1 coefficients: were swept over $[10^{-11}, 1]$) for optimal sparsity reconstruction tradeoff.
    \item Auxiliary Loss: Ghost gradients were added to prevent dead features.
\end{itemize}

For a full list of the SAEs, see Table \ref{tab:models_structured}, and for evaluation metrics, see Appendix \ref{section:saetrainingddetails}.




\subsection{Open Source Toy Vision Transformers}
\label{sec:toy_models}
We release toy \glspl{vit} trained on ImageNet with 1–4 layers (\cref{appendix:imagenet_toy_models}). These models include both full transformer architectures and attention-only variants, with multiple training checkpoints provided. Toy models enable researchers to validate hypotheses in a controlled environment before scaling to full-size networks—a strategy that has proven valuable in language mechanistic interpretability \cite{elhage2022toy, olsson2022context, nanda2023progress}.



\section{Preliminary SAE Analysis with Prisma}

We use the Prisma toolkit to briefly present two unexpected observations about CLIP SAEs: \textbf{1)} The internal representations of CLIP \glspl{vit} appear much less sparse than in language, leaving optimal sparsity for vision SAEs an open question. \textbf{2)} There is sometimes a \textit{decrease} in the model's cross-entropy loss when injecting SAE reconstructions into the forward pass, which may be explained by the denoising properties of SAEs. We present our analysis here and leave further exploration to the broader vision mechanistic interpretability community.

\subsection{Vision SAE Sparsity is Higher than Language SAE Sparsity}
High-quality SAEs balance low sparsity, measure by L0 (number of active SAE latents per token), with high explained variance. For language models like GPT-2 Small, the L0 ranges from 12–74 per token \cite{bloom2024gpt2residualsaes}, with top-K SAEs performing best at K = 32–256 \cite{gao2024scaling}.

Vision SAEs exhibit significantly higher L0~\cite{joseph2025steeringclipsvisiontransformer}: approximately 500+ per patch in CLIP-B/32 while retaining a comparable explained variance to language (see  SAE evaluation tables in \cref{section: sae_evaluation_details}). Even with improved training techniques, our top-K SAEs (K = 64, 128) still show polysemantic features, suggesting higher K values may be necessary. Analysis of a pretrained vision SAE \cite{daujotas2024interpreting} from the Kandinsky encoder reveals an extreme L0 of 122,631 per CLS token, despite its demonstrated effectiveness in diffusion model steering.

\begin{figure}
    \centering
    \includegraphics[width=\linewidth]{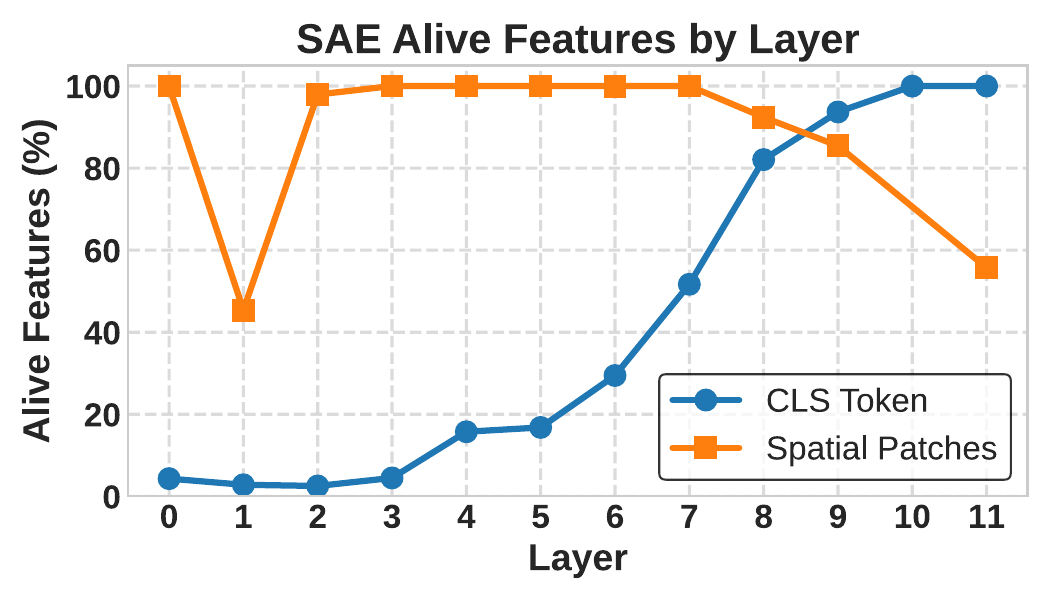}
    \caption{The percentage of alive features shows that CLS tokens require more features in deeper layers while spatial tokens show declining utilization, showing how information shifts from spatial patches to a global representation. Results are from the vanilla CLS SAEs (\cref{table:cls_only}).}
    \label{fig:alive_features}
\end{figure}

We propose four potential explanations for this disparity:

\begin{enumerate}
\item \emph{Information Density Differences.} Visual inputs may have an inherently denser distribution than language, lack the power-law distribution of natural language, and may require more extensive preprocessing.

\item \emph{Patch Granularity.} CLIP-B/32 processes approximately 50 patches compared to GPT-2's 1024 tokens, a 20× difference. When normalized (25 per patch), vision L0 becomes more comparable to language. The relationship between patch size and L0 warrants further investigation.

\item \emph{Patch Type Specialization.} Spatial patches and CLS tokens exhibit distinct distributional properties \cite{vilas2023analyzing, lim2024sparse, joseph2025steeringclipsvisiontransformer}. The CLS token's high L0 encapsulates a global image representation. When normalized (L0 $\approx 479$ per patch) and adjusted to language-scale tokenization (L0 $\approx 120$), this aligns with language SAE sparsity. Our analysis of dead SAE features reveals complementary processing patterns: the CLS token has less dead features in deeper layers as it builds a global representation, while spatial patches show an inverse trend as information moves to the CLS token (Fig. \ref{fig:alive_features}). This suggests optimal top-K varies by token type and layer depth, with CLS tokens requiring higher K in later layers to accommodate their enriched representations.

\item \emph{Domain-Specific Optimization.} Current vision SAE training methods are adapted from language mechanistic interpretability, potentially limiting achievable sparsity. Vision-specific SAE training techniques may yield sparser representations, but are yet to be developed.
\end{enumerate}

This apparent sparsity discrepancy across modalities remains an open question, and we hope these insights serve as a starting point for further research.

\begin{figure}
    \centering
    \includegraphics[width=\linewidth]{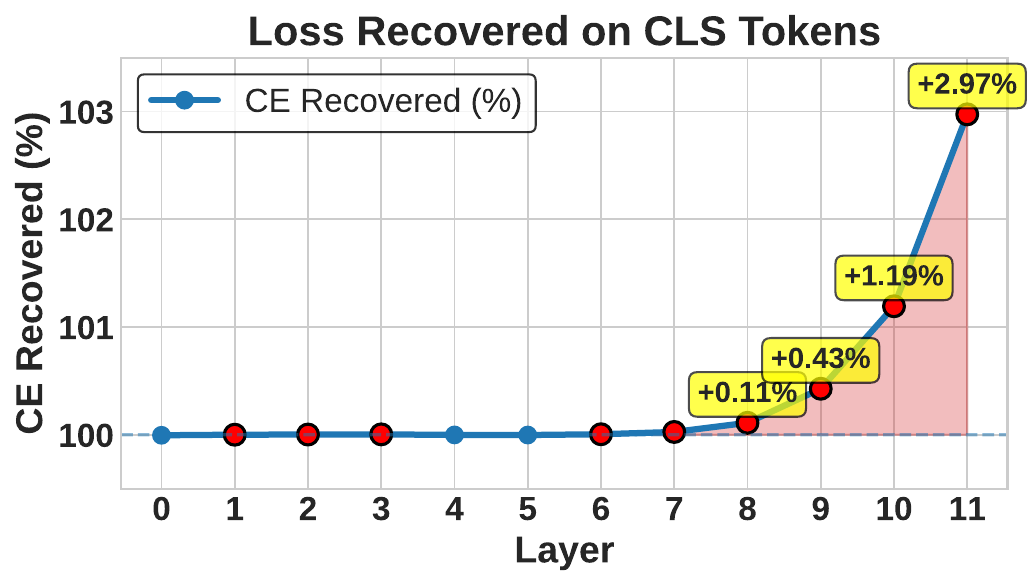}
    \caption{Using an SAE during the model's forward pass \textit{decreases} the original model's loss, with results becoming more dramatic in later layers. Red dots indicate for which layers the performance is improved. Results are from the CLIP-B Top K = 64 SAEs (\cref{table:top_k_cls_64}).}
    \label{fig:sae_performance}
\end{figure}

\subsection{\gls{sae} Reconstructions Can Improve Model Loss}
Our analysis reveals the unexpected finding that inserting \gls{sae} reconstructions into the forward pass can reduce cross-entropy loss in the original model. In particular, the Vanilla and Top K = 64 CLS-only CLIP SAE sets (\cref{section: sae_evaluation_details}, \cref{table:vanilla_all_patches} and \cref{table:top_k_cls_64}), and the vanilla DINO set (\cref{table:dino_vanilla}) have layers showing this phenomenon. For the Top K = 64 CLS-only SAEs, the effect intensifies as the CLS token representations develop through network layers (\cref{fig:sae_performance}), suggesting a potential denoising mechanism at work, and aligning with a previous observation of SAE reconstructions reducing the loss in CLIP's final layer \cite{fry2024mulitmodal}. Interestingly, we do not observe significant improvements in the CLIP transcoders (\cref{tab:clip_transcoders}), CLIP Vanilla Spatial Patch SAEs (\cref{sec:spatial_patches}), or CLIP Vanilla CLS-only SAEs (\cref{table:cls_only}). The precise mechanism behind this improvement remains unclear and presents a compelling avenue for future research by the vision mechanistic interpretability community. 

\section{Conclusion}
Prisma bridges language and vision interpretability techniques with an extensive toolkit including 80+ SAEs for CLIP, DINO, and video transformers, and experimental toy models. Our analysis reveals vision representations are less sparse than language models, and SAE reconstructions can improve model performance, especially for CLS tokens in deeper layers. We invite researchers to build on these findings and contribute to Prisma's open-source development

\section{Acknowledgements}
We thank Joseph Bloom, Neel Nanda, Jacob Dunefsky, and Leo Gao for the discussions during the early days of the repository. 

{
    \small
    \bibliographystyle{ieeenat_fullname}
    \bibliography{main}

\begin{thebibliography}{66}
\providecommand{\natexlab}[1]{#1}
\providecommand{\url}[1]{\texttt{#1}}
\expandafter\ifx\csname urlstyle\endcsname\relax
  \providecommand{\doi}[1]{doi: #1}\else
  \providecommand{\doi}{doi: \begingroup \urlstyle{rm}\Url}\fi

\bibitem[Achtibat et~al.(2023)Achtibat, Dreyer, Eisenbraun, Bosse, Wiegand, Samek, and Lapuschkin]{achtibat2023attribution}
Reduan Achtibat, Maximilian Dreyer, Ilona Eisenbraun, Sebastian Bosse, Thomas Wiegand, Wojciech Samek, and Sebastian Lapuschkin.
\newblock From attribution maps to human-understandable explanations through concept relevance propagation.
\newblock \emph{Nature Machine Intelligence}, 5\penalty0 (9):\penalty0 1006--1019, 2023.

\bibitem[Arnab et~al.(2021)Arnab, Dehghani, Heigold, Sun, Lucic, and Schmid]{vivit2021}
Anurag Arnab, Mostafa Dehghani, Georg Heigold, Chen Sun, Mario Lucic, and Cordelia Schmid.
\newblock Vivit: {A} video vision transformer.
\newblock \emph{CoRR}, abs/2103.15691, 2021.

\bibitem[Bardes et~al.(2024)Bardes, Garrido, Ponce, Chen, Rabbat, LeCun, Assran, and Ballas]{bardes2024revisitingfeaturepredictionlearning}
Adrien Bardes, Quentin Garrido, Jean Ponce, Xinlei Chen, Michael Rabbat, Yann LeCun, Mahmoud Assran, and Nicolas Ballas.
\newblock Revisiting feature prediction for learning visual representations from video, 2024.

\bibitem[Bloom(2024)]{bloom2024gpt2residualsaes}
Joseph Bloom.
\newblock Open source sparse autoencoders for all residual stream layers of gpt2 small.
\newblock \url{https://www.alignmentforum.org/posts/f9EgfLSurAiqRJySD/open-source-sparse-autoencoders-for-all-residual-stream}, 2024.

\bibitem[Bricken et~al.(2023)Bricken, Templeton, Batson, Chen, Jermyn, Conerly, Turner, Anil, Denison, Askell, Lasenby, Wu, Kravec, Schiefer, Maxwell, Joseph, Hatfield-Dodds, Tamkin, Nguyen, McLean, Burke, Hume, Carter, Henighan, and Olah]{bricken2023monosemanticity}
Trenton Bricken, Adly Templeton, Joshua Batson, Brian Chen, Adam Jermyn, Thomas Conerly, Nick Turner, Cem Anil, Catherine Denison, Amanda Askell, Richard Lasenby, Yuntao Wu, Stephen Kravec, Noa Schiefer, Thomas Maxwell, Nicholas Joseph, Zac Hatfield-Dodds, Alex Tamkin, Katy Nguyen, Brayden McLean, James~E. Burke, Thomas Hume, Shan Carter, Tom Henighan, and Christopher Olah.
\newblock Towards monosemanticity: Decomposing language models with dictionary learning.
\newblock \emph{Transformer Circuits Thread}, 2023.

\bibitem[Caron et~al.(2021)Caron, Touvron, Misra, J{\'e}gou, Mairal, Bojanowski, and Joulin]{caron2021emerging}
Mathilde Caron, Hugo Touvron, Ishan Misra, Herv{\'e} J{\'e}gou, Julien Mairal, Piotr Bojanowski, and Armand Joulin.
\newblock Emerging properties in self-supervised vision transformers.
\newblock In \emph{Proceedings of the IEEE/CVF International Conference on Computer Vision}, pages 9650--9660, 2021.

\bibitem[Choi et~al.(2019)Choi, Gao, Messou, and Huang]{choi2019cantidancemall}
Jinwoo Choi, Chen Gao, Joseph C.~E. Messou, and Jia-Bin Huang.
\newblock Why can't i dance in the mall? learning to mitigate scene bias in action recognition, 2019.

\bibitem[Conerly et~al.(2025)Conerly, Cunningham, Templeton, Lindsey, Hosmer, and Jermyn]{DictionaryOptimisation2023Conerly}
Tom Conerly, Hoagy Cunningham, Adly Templeton, Jack Lindsey, Basil Hosmer, and Adam Jermyn.
\newblock Dictionary learning optimization techniques, 2025.
\newblock Accessed: 2025-03-10.

\bibitem[Conmy et~al.(2023)Conmy, Mavor-Parker, Lynch, Heimersheim, and Garriga-Alonso]{conmy2023towards}
Arthur Conmy, Augustine~N Mavor-Parker, Aengus Lynch, Stefan Heimersheim, and Adri{\`a} Garriga-Alonso.
\newblock Towards automated circuit discovery for mechanistic interpretability.
\newblock \emph{arXiv preprint arXiv:2304.14997}, 2023.

\bibitem[Cooney and Nanda(2023)]{cooney2023circuitsvis}
Alan Cooney and Neel Nanda.
\newblock Circuitsvis.
\newblock \url{https://github.com/TransformerLensOrg/CircuitsVis}, 2023.

\bibitem[Cunningham et~al.(2023)Cunningham, Ewart, Smith, Huben, and Sharkey]{cunningham2023sparse}
Hoagy Cunningham, Arthur Ewart, Lewis Smith, Robert Huben, and Lee Sharkey.
\newblock Sparse autoencoders find highly interpretable model directions.
\newblock \emph{arXiv preprint arXiv:2309.08600}, 2023.

\bibitem[Daujotas(2024)]{daujotas2024interpreting}
Gytis Daujotas.
\newblock Interpreting and steering features in images.
\newblock \emph{URL https://www. lesswrong. com/posts/Quqekpvx8BGMMcaem/interpreting-and-steering-features-in-images}, 2024.

\bibitem[Deng et~al.(2009)Deng, Dong, Socher, Li, Li, and Fei-Fei]{deng2009imagenet}
Jia Deng, Wei Dong, Richard Socher, Li-Jia Li, Kai Li, and Li Fei-Fei.
\newblock Imagenet: A large-scale hierarchical image database.
\newblock In \emph{2009 IEEE conference on computer vision and pattern recognition}, pages 248--255. Ieee, 2009.

\bibitem[Dosovitskiy et~al.(2020)Dosovitskiy, Beyer, Kolesnikov, Weissenborn, Zhai, Unterthiner, Dehghani, Minderer, Heigold, Gelly, et~al.]{dosovitskiy2020image}
Alexey Dosovitskiy, Lucas Beyer, Alexander Kolesnikov, Dirk Weissenborn, Xiaohua Zhai, Thomas Unterthiner, Mostafa Dehghani, Matthias Minderer, Georg Heigold, Sylvain Gelly, et~al.
\newblock An image is worth 16x16 words: Transformers for image recognition at scale.
\newblock \emph{arXiv preprint arXiv:2010.11929}, 2020.

\bibitem[Dunefsky et~al.(2024)Dunefsky, Chlenski, and Nanda]{dunefsky2024transcoders}
Jacob Dunefsky, Peter Chlenski, and Neel Nanda.
\newblock Transcoders enable fine-grained interpretable circuit analysis for language models.
\newblock 2024.

\bibitem[Dunefsky et~al.(2025)Dunefsky, Chlenski, and Nanda]{dunefsky2025transcoders}
Jacob Dunefsky, Philippe Chlenski, and Neel Nanda.
\newblock Transcoders find interpretable llm feature circuits.
\newblock \emph{Advances in Neural Information Processing Systems}, 37:\penalty0 24375--24410, 2025.

\bibitem[EleutherAI(2024{\natexlab{a}})]{eleutherSparsify}
EleutherAI.
\newblock Sparsify.
\newblock GitHub, 2024{\natexlab{a}}.
\newblock Accessed: 2025-03-10.

\bibitem[EleutherAI(2024{\natexlab{b}})]{eleutherai2024sparsify}
EleutherAI.
\newblock Sparsify: Sparsify transformers with saes and transcoders, 2024{\natexlab{b}}.

\bibitem[Elhage et~al.(2022{\natexlab{a}})Elhage, Hume, Olsson, Schiefer, Henighan, Kravec, Hatfield-Dodds, Lasenby, Drain, Chen, Grosse, McCandlish, Kaplan, Amodei, Wattenberg, and Olah]{elhage2022toymodelssuperposition}
Nelson Elhage, Tristan Hume, Catherine Olsson, Nicholas Schiefer, Tom Henighan, Shauna Kravec, Zac Hatfield-Dodds, Robert Lasenby, Dawn Drain, Carol Chen, Roger Grosse, Sam McCandlish, Jared Kaplan, Dario Amodei, Martin Wattenberg, and Christopher Olah.
\newblock Toy models of superposition, 2022{\natexlab{a}}.

\bibitem[Elhage et~al.(2022{\natexlab{b}})Elhage, Hume, Olsson, Schiefer, Henighan, Kravec, Hatfield-Dodds, Lasenby, Drain, Chen, et~al.]{elhage2022toy}
Nelson Elhage, Tristan Hume, Catherine Olsson, Nicholas Schiefer, Tom Henighan, Shauna Kravec, Zac Hatfield-Dodds, Robert Lasenby, Dawn Drain, Carol Chen, et~al.
\newblock Toy models of superposition.
\newblock \emph{arXiv preprint arXiv:2209.10652}, 2022{\natexlab{b}}.

\bibitem[Fiotto-Kaufman et~al.(2024)Fiotto-Kaufman, Loftus, Todd, Brinkmann, Juang, Pal, Rager, Mueller, Marks, Sharma, Lucchetti, Ripa, Belfki, Prakash, Multani, Brodley, Guha, Bell, Wallace, and Bau]{fiottokaufman2024nnsightndifdemocratizingaccess}
Jaden Fiotto-Kaufman, Alexander~R Loftus, Eric Todd, Jannik Brinkmann, Caden Juang, Koyena Pal, Can Rager, Aaron Mueller, Samuel Marks, Arnab~Sen Sharma, Francesca Lucchetti, Michael Ripa, Adam Belfki, Nikhil Prakash, Sumeet Multani, Carla Brodley, Arjun Guha, Jonathan Bell, Byron Wallace, and David Bau.
\newblock Nnsight and ndif: Democratizing access to foundation model internals.
\newblock 2024.

\bibitem[Fry(2024)]{fry2024mulitmodal}
Hugo Fry.
\newblock Towards multimodal interpretability: Learning sparse interpretable features in vision transformers.
\newblock \url{https://www.lesswrong.com/posts/bCtbuWraqYTDtuARg/towards-multimodal-interpretability-learning-sparse-2}, 2024.

\bibitem[Gandelsman et~al.(2024{\natexlab{a}})Gandelsman, Efros, and Steinhardt]{gandelsman2024interpretingsecondordereffectsneurons}
Yossi Gandelsman, Alexei~A. Efros, and Jacob Steinhardt.
\newblock Interpreting the second-order effects of neurons in clip, 2024{\natexlab{a}}.

\bibitem[Gandelsman et~al.(2024{\natexlab{b}})Gandelsman, Efros, and Steinhardt]{gandelsmanclipdecomposition}
Yossi Gandelsman, Alexei~A. Efros, and Jacob Steinhardt.
\newblock Interpreting clip's image representation via text-based decomposition, 2024{\natexlab{b}}.

\bibitem[Gao et~al.(2024{\natexlab{a}})Gao, la~Tour, Tillman, Goh, Troll, Radford, Sutskever, Leike, and Wu]{gao2024scaling}
Leo Gao, Tom~Dupr{\'e} la Tour, Henk Tillman, Gabriel Goh, Rajan Troll, Alec Radford, Ilya Sutskever, Jan Leike, and Jeffrey Wu.
\newblock Scaling and evaluating sparse autoencoders.
\newblock \emph{arXiv preprint arXiv:2406.04093}, 2024{\natexlab{a}}.

\bibitem[Gao et~al.(2024{\natexlab{b}})Gao, la~Tour, Tillman, Goh, Troll, Radford, Sutskever, Leike, and Wu]{gao2024scalingevaluatingsparseautoencoders}
Leo Gao, Tom~Dupré la Tour, Henk Tillman, Gabriel Goh, Rajan Troll, Alec Radford, Ilya Sutskever, Jan Leike, and Jeffrey Wu.
\newblock Scaling and evaluating sparse autoencoders, 2024{\natexlab{b}}.

\bibitem[Ghodrati et~al.(2018)Ghodrati, Gavves, and Snoek]{DBLP:journals/corr/abs-1807-06980}
Amir Ghodrati, Efstratios Gavves, and Cees G.~M. Snoek.
\newblock Video time: Properties, encoders and evaluation.
\newblock \emph{CoRR}, abs/1807.06980, 2018.

\bibitem[Heimersheim and Nanda(2024)]{heimersheim2024useinterpretactivationpatching}
Stefan Heimersheim and Neel Nanda.
\newblock How to use and interpret activation patching, 2024.

\bibitem[Ilharco et~al.(2021)Ilharco, Wortsman, Wightman, Gordon, Carlini, Taori, Dave, Shankar, Namkoong, Miller, Hajishirzi, Farhadi, and Schmidt]{ilharco_gabriel_2021_5143773}
Gabriel Ilharco, Mitchell Wortsman, Ross Wightman, Cade Gordon, Nicholas Carlini, Rohan Taori, Achal Dave, Vaishaal Shankar, Hongseok Namkoong, John Miller, Hannaneh Hajishirzi, Ali Farhadi, and Ludwig Schmidt.
\newblock Openclip, 2021.

\bibitem[Ilic et~al.(2022)Ilic, Pock, and Wildes]{ilic2022appearancefreeactionrecognition}
Filip Ilic, Thomas Pock, and Richard~P. Wildes.
\newblock Is appearance free action recognition possible?, 2022.

\bibitem[Jermyn and Templeton(2024)]{GhostGrads2023JermynTempleton}
A Jermyn and A Templeton.
\newblock Ghost grads: An improvement on resampling, 2024.
\newblock Accessed: 2025-03-10.

\bibitem[Joseph(2024)]{joseph2024vitprismanotebook}
Sonia Joseph.
\newblock Vit-prisma main demo.
\newblock Google Colab Jupyter Notebook, 2024.
\newblock Accessed: March 10, 2025.

\bibitem[Joseph and Nanda(2024)]{joseph2024laying}
Sonia Joseph and Neel Nanda.
\newblock Laying the foundations for vision and multimodal mechanistic interpretability \& open problems.
\newblock \emph{AI Alignment Forum}, 2024.

\bibitem[Joseph et~al.(2023)Joseph, Zholus, Samsami, and Richards]{joseph2023mining}
S. Joseph, A. Zholus, M.~R. Samsami, and B.~A. Richards.
\newblock Mining the diamond miner: Mechanistic interpretability on the video pretraining agent.
\newblock ATTRIB Workshop at NeurIPS 2023, 2023.
\newblock Available at: \url{https://openreview.net/pdf?id=lDeysxpH6W}.

\bibitem[Joseph et~al.(2025)Joseph, Suresh, Goldfarb, Hufe, Gandelsman, Graham, Bzdok, Samek, and Richards]{joseph2025steeringclipsvisiontransformer}
Sonia Joseph, Praneet Suresh, Ethan Goldfarb, Lorenz Hufe, Yossi Gandelsman, Robert Graham, Danilo Bzdok, Wojciech Samek, and Blake~Aaron Richards.
\newblock Steering clip's vision transformer with sparse autoencoders, 2025.

\bibitem[Joseph~Bloom and Chanin(2024)]{bloom2024saetrainingcodebase}
Curt~Tigges Joseph~Bloom and David Chanin.
\newblock Saelens.
\newblock \url{https://github.com/jbloomAus/SAELens}, 2024.

\bibitem[Jucys et~al.(2024)Jucys, Adamopoulos, Hamidi, Milani, Samsami, Zholus, Joseph, Richards, Rish, and {\c{S}}im{\c{s}}ek]{jucys2024interpretability}
Karolis Jucys, George Adamopoulos, Mehrab Hamidi, Stephanie Milani, Mohammad~Reza Samsami, Artem Zholus, Sonia Joseph, Blake Richards, Irina Rish, and {\"O}zg{\"u}r {\c{S}}im{\c{s}}ek.
\newblock Interpretability in action: Exploratory analysis of vpt, a minecraft agent.
\newblock \emph{arXiv preprint arXiv:2407.12161}, 2024.

\bibitem[Kingma and Ba(2014)]{kingma2014adam}
Diederik~P Kingma and Jimmy Ba.
\newblock Adam: A method for stochastic optimization.
\newblock \emph{arXiv preprint arXiv:1412.6980}, 2014.

\bibitem[Komorowski et~al.(2023)Komorowski, Baniecki, and Biecek]{komorowski2023towards}
Piotr Komorowski, Hubert Baniecki, and Przemyslaw Biecek.
\newblock Towards evaluating explanations of vision transformers for medical imaging.
\newblock In \emph{Proceedings of the IEEE/CVF conference on computer vision and pattern recognition}, pages 3726--3732, 2023.

\bibitem[Kowal et~al.(2024{\natexlab{a}})Kowal, Dave, Ambrus, Gaidon, Derpanis, and Tokmakov]{Kowal_2024_CVPR}
Matthew Kowal, Achal Dave, Rares Ambrus, Adrien Gaidon, Konstantinos~G. Derpanis, and Pavel Tokmakov.
\newblock Understanding video transformers via universal concept discovery.
\newblock In \emph{Proceedings of the IEEE/CVF Conference on Computer Vision and Pattern Recognition (CVPR)}, pages 10946--10956, 2024{\natexlab{a}}.

\bibitem[Kowal et~al.(2024{\natexlab{b}})Kowal, Siam, Islam, Bruce, Wildes, and Derpanis]{kowal2024quantifyinglearningstaticvs}
Matthew Kowal, Mennatullah Siam, Md~Amirul Islam, Neil D.~B. Bruce, Richard~P. Wildes, and Konstantinos~G. Derpanis.
\newblock Quantifying and learning static vs. dynamic information in deep spatiotemporal networks, 2024{\natexlab{b}}.

\bibitem[Kramár et~al.(2024)Kramár, Lieberum, Shah, and Nanda]{kramar2024atpefficientscalablemethod}
János Kramár, Tom Lieberum, Rohin Shah, and Neel Nanda.
\newblock Atp*: An efficient and scalable method for localizing llm behaviour to components, 2024.

\bibitem[Li and Vasconcelos(2019)]{li2019repairremovingrepresentationbias}
Yi Li and Nuno Vasconcelos.
\newblock Repair: Removing representation bias by dataset resampling, 2019.

\bibitem[Li et~al.(2018)Li, Li, and Vasconcelos]{Li_2018_ECCV}
Yingwei Li, Yi Li, and Nuno Vasconcelos.
\newblock Resound: Towards action recognition without representation bias.
\newblock In \emph{Proceedings of the European Conference on Computer Vision (ECCV)}, 2018.

\bibitem[Lieberum et~al.(2024)Lieberum, Rajamanoharan, Conmy, Smith, Sonnerat, Varma, Kramár, Dragan, Shah, and Nanda]{lieberum2024gemmascopeopensparse}
Tom Lieberum, Senthooran Rajamanoharan, Arthur Conmy, Lewis Smith, Nicolas Sonnerat, Vikrant Varma, János Kramár, Anca Dragan, Rohin Shah, and Neel Nanda.
\newblock Gemma scope: Open sparse autoencoders everywhere all at once on gemma 2, 2024.

\bibitem[Lim et~al.(2024)Lim, Choi, Choo, and Schneider]{lim2024sparse}
Hyesu Lim, Jinho Choi, Jaegul Choo, and Steffen Schneider.
\newblock Sparse autoencoders reveal selective remapping of visual concepts during adaptation.
\newblock \emph{arXiv preprint arXiv:2412.05276}, 2024.

\bibitem[Lindsey et~al.(2024)Lindsey, Templeton, Marcus, Conerly, Batson, and Olah]{lindsey2024sparse}
Jack Lindsey, Adly Templeton, Jonathan Marcus, Thomas Conerly, Joshua Batson, and Christopher Olah.
\newblock Sparse crosscoders for cross-layer features and model diffing.
\newblock \emph{Transformer Circuits Thread}, 2024.
\newblock Research Update.

\bibitem[Marks(2023)]{OpenSourceDictionaries2023Marks}
Sam Marks.
\newblock Some open-source dictionaries and dictionary learning infrastructure, 2023.
\newblock Accessed: 2025-03-10.

\bibitem[Marks(2024)]{marks2024dictionary}
Samuel Marks.
\newblock dictionary\_learning.
\newblock 2024.

\bibitem[Marks et~al.(2024)Marks, Rager, Michaud, Belinkov, Bau, and Mueller]{marks2024sparsefeaturecircuitsdiscovering}
Samuel Marks, Can Rager, Eric~J. Michaud, Yonatan Belinkov, David Bau, and Aaron Mueller.
\newblock Sparse feature circuits: Discovering and editing interpretable causal graphs in language models, 2024.

\bibitem[Nanda(2023)]{nanda2023attributionpatching}
Neel Nanda.
\newblock Attribution patching: Activation patching at industrial scale, 2023.

\bibitem[Nanda and Bloom(2022)]{nanda2022transformerlens}
Neel Nanda and Joseph Bloom.
\newblock Transformerlens.
\newblock \url{https://github.com/TransformerLensOrg/TransformerLens}, 2022.

\bibitem[Nanda et~al.(2023)Nanda, Chan, Lieberum, Smith, and Steinhardt]{nanda2023progress}
Neel Nanda, Lawrence Chan, Tom Lieberum, Jess Smith, and Jacob Steinhardt.
\newblock Progress measures for grokking via mechanistic interpretability.
\newblock \emph{arXiv preprint arXiv:2301.05217}, 2023.

\bibitem[Naseer et~al.(2021)Naseer, Ranasinghe, Khan, Khan, and Porikli]{naseer2021intriguing}
Muzammal Naseer, Kanchana Ranasinghe, Salman Khan, Fahad~Shahbaz Khan, and Fatih Porikli.
\newblock Intriguing properties of vision transformers.
\newblock \emph{Advances in Neural Information Processing Systems}, 34:\penalty0 23296--23308, 2021.

\bibitem[nostalgebraist(2020)]{nostalgebraist2020logitlens}
nostalgebraist.
\newblock Interpreting gpt: The logit lens, 2020.
\newblock LessWrong.

\bibitem[Olsson et~al.(2022)Olsson, Elhage, Nanda, Joseph, DasSarma, Henighan, Mann, Askell, Bai, Chen, et~al.]{olsson2022context}
Catherine Olsson, Nelson Elhage, Neel Nanda, Nicholas Joseph, Nova DasSarma, Tom Henighan, Ben Mann, Amanda Askell, Yuntao Bai, Anna Chen, et~al.
\newblock In-context learning and induction heads.
\newblock \emph{arXiv preprint arXiv:2209.11895}, 2022.

\bibitem[Oquab et~al.(2023)Oquab, Darcet, Moutakanni, Vo, Szafraniec, Khalidov, Fernandez, Haziza, Massa, El-Nouby, et~al.]{oquab2023dinov2}
Maxime Oquab, Timoth{\'e}e Darcet, Th{\'e}o Moutakanni, Huy Vo, Marc Szafraniec, Vasil Khalidov, Pierre Fernandez, Daniel Haziza, Francisco Massa, Alaaeldin El-Nouby, et~al.
\newblock Dinov2: Learning robust visual features without supervision.
\newblock \emph{arXiv preprint arXiv:2304.07193}, 2023.

\bibitem[Radford et~al.(2021)Radford, Kim, Hallacy, Ramesh, Goh, Agarwal, Sastry, Askell, Mishkin, Clark, et~al.]{radford2021learning}
Alec Radford, Jong~Wook Kim, Chris Hallacy, Aditya Ramesh, Gabriel Goh, Sandhini Agarwal, Girish Sastry, Amanda Askell, Pamela Mishkin, Jack Clark, et~al.
\newblock Learning transferable visual models from natural language supervision.
\newblock \emph{arXiv preprint arXiv:2103.00020}, 2021.

\bibitem[Raghu et~al.(2021)Raghu, Unterthiner, Kornblith, Zhang, and Dosovitskiy]{raghu2021vision}
Maithra Raghu, Thomas Unterthiner, Simon Kornblith, Chiyuan Zhang, and Alexey Dosovitskiy.
\newblock Do vision transformers see like convolutional neural networks?
\newblock \emph{Advances in Neural Information Processing Systems}, 34:\penalty0 12116--12128, 2021.

\bibitem[Rajamanoharan et~al.(2024{\natexlab{a}})Rajamanoharan, Conmy, Smith, Lieberum, Varma, Kramár, Shah, and Nanda]{rajamanoharan2024improvingdictionarylearninggated}
Senthooran Rajamanoharan, Arthur Conmy, Lewis Smith, Tom Lieberum, Vikrant Varma, János Kramár, Rohin Shah, and Neel Nanda.
\newblock Improving dictionary learning with gated sparse autoencoders, 2024{\natexlab{a}}.

\bibitem[Rajamanoharan et~al.(2024{\natexlab{b}})Rajamanoharan, Lieberum, Sonnerat, Conmy, Varma, Kramár, and Nanda]{rajamanoharan2024jumpingaheadimprovingreconstruction}
Senthooran Rajamanoharan, Tom Lieberum, Nicolas Sonnerat, Arthur Conmy, Vikrant Varma, János Kramár, and Neel Nanda.
\newblock Jumping ahead: Improving reconstruction fidelity with jumprelu sparse autoencoders, 2024{\natexlab{b}}.

\bibitem[Rajaram et~al.(2024)Rajaram, Chowdhury, Torralba, Andreas, and Schwettmann]{rajaram2024automaticdiscoveryvisualcircuits}
Achyuta Rajaram, Neil Chowdhury, Antonio Torralba, Jacob Andreas, and Sarah Schwettmann.
\newblock Automatic discovery of visual circuits, 2024.

\bibitem[Templeton et~al.(2024{\natexlab{a}})Templeton, Batson, Jermyn, and Olah]{templeton2024predicting}
Adly Templeton, Joshua Batson, Adam Jermyn, and Christopher Olah.
\newblock Predicting future activations.
\newblock 2024{\natexlab{a}}.

\bibitem[Templeton et~al.(2024{\natexlab{b}})Templeton, Conerly, Marcus, Henighan, Golubeva, and Bricken]{ImprovementDictionary2023Templeton}
Adly Templeton, Tom Conerly, Jonathan Marcus, Tom Henighan, Anna Golubeva, and Trenton Bricken.
\newblock Improvements to dictionary learning, 2024{\natexlab{b}}.
\newblock Accessed: 2025-03-10.

\bibitem[Vilas et~al.(2023)Vilas, Schauml{\"o}ffel, and Roig]{vilas2023analyzing}
Martina~G Vilas, Timothy Schauml{\"o}ffel, and Gemma Roig.
\newblock Analyzing vision transformers for image classification in class embedding space.
\newblock \emph{Advances in neural information processing systems}, 36:\penalty0 40030--40041, 2023.

\bibitem[Wang et~al.(2022)Wang, Variengien, Conmy, Shlegeris, and Steinhardt]{wang2022interpretability}
Kevin Wang, Alexandre Variengien, Arthur Conmy, Buck Shlegeris, and Jacob Steinhardt.
\newblock Interpretability in the wild: a circuit for indirect object identification in gpt-2 small.
\newblock \emph{arXiv preprint arXiv:2211.00593}, 2022.

\end{thebibliography}
}

\clearpage
\setcounter{page}{1}
\maketitlesupplementary

\section{Open Source SAE Suite List}

We train sparse coders on all twelve layers of the residual stream for CLIP-B-32 and DINO-B-32 in the following kinds:

\begin{table}[h]
\centering
\begin{tabular}{l l l l l}
\toprule
\textbf{Model} & \textbf{Coder} & \textbf{Version} & \textbf{Token} \\
\midrule
CLIP-B   & SAE         & ReLU  & All   \\
\midrule
CLIP-B   & SAE         & ReLU  & CLS     \\
CLIP-B   & SAE         & Top K (K = 64) & CLS   \\
\midrule
CLIP-B   & SAE         & ReLU  & Spatial  \\
CLIP-B   & SAE         & Top K (K = 64) & Spatial   \\
CLIP-B   & SAE         & Top K (K = 128) & Spatial \\
\midrule
DINO-B   & SAE         & ReLU  & All  \\
\midrule
CLIP     & Transcoder  & ReLU  & All  \\
\bottomrule
\end{tabular}
\caption{Open Source SAE and Transcoder Weights for All Layers}
\label{tab:models_structured}
\end{table}

\section{Details of Toy Vision Transformer Models}
\label{appendix:toy_models_details}

\label{appendix:toy_models}

\subsection{ImageNet}
\label{appendix:imagenet_toy_models}

Our primary toy \glspl{vit} use a patch size of 16. For enhanced attention-head visualization, we include a subset with a patch size of 32 (see Appendix \ref{appendix:imagenet_toy_models}). The attention-only variants are more amenable to mathematical analysis than full transformer architectures.

\subsubsection{Patch Size 16}

\begin{table}[h]
\centering
\caption{Accuracy [Top-1 $\vert$ Top-5]}
\begin{tabular}{lllll}
\hline
\textbf{Size} & \textbf{Num Layers} & \textbf{Attn+MLP} & \textbf{Attn-Only} \\
\hline
tiny & 1 & 0.16 $\vert$ 0.33 & 0.11 $\vert$ 0.25 \\
base & 2 & 0.23 $\vert$ 0.44 & 0.16 $\vert$ 0.34 \\
small & 3 & 0.28 $\vert$ 0.51 & 0.17 $\vert$ 0.35  \\
medium & 4 & 0.33 $\vert$ 0.56 & 0.17 $\vert$ 0.36  \\
\hline
\end{tabular}
\label{table:toy_vit}
\end{table}

\newpage

\subsubsection{Patch Size 32}

We trained a toy ViT with a larger patch size ViT (32), allowing for inspectable attention heads by the attention head visualizer, whereas patch size 16 attention heads are too large to easily render in JavaScript. This model includes fifty training checkpoints for analyzing training dynamics.

 \begin{table}[!h]
\centering
\begin{tabular}{lllll}
\hline
\textbf{Size} & \textbf{Num Layers} & \textbf{Attn+MLP} & \textbf{Attn-Only} \\
\hline
tiny & 3 & 0.22 $\vert$ 0.42 & N/A \\
\hline
\end{tabular}
\caption{ImageNet-1k Classification Checkpoints (patch size 32) [Top-1 $\vert$ Top-5].}
\label{tab:checkpoints}
\end{table}







\clearpage
\onecolumn

\section{SAE Evaluations}
\label{section: sae_evaluation_details}

Below are our SAE evaluations from the SAEs that we trained in \cref{section:saetrainingddetails}.

\subsection{CLIP-ViT-B-32 SAEs}

\subsubsection*{Vanilla SAEs (All Patches)}
\captionof{table}{CLIP-ViT-B-32 vanilla sparse autoencoder performance metrics for all patches (CLS + spatial).}
\label{table:vanilla_all_patches}
\footnotesize
\resizebox{\textwidth}{!}{%
\begin{tabular}{|c|c|c|c|c|c|c|c|c|c|c|c|c|l|}
\hline
\textbf{Layer} & \textbf{Sublayer} & \textbf{l1 coeff.} & \textbf{\% Explained var.} & \textbf{Avg L0} & \textbf{Avg CLS L0} & \textbf{Cos sim} & \textbf{Recon cos sim} & \textbf{CE} & \textbf{Recon CE} & \textbf{Zero abl CE} & \textbf{\% CE recovered} & \textbf{\% Alive features} & \textbf{Model} \\
\hline
0 & mlp\_out    & 1e-5  & 98.7 & 604.44  & 36.92   & 0.994 & 0.998 & 6.762 & 6.762 & 6.779 & 99.51  & 100    & \href{https://huggingface.co/prisma-multimodal/sparse-autoencoder-clip-b-32-sae-vanilla-x64-layer-0-hook_mlp_out-l1-1e-05}{link} \\
0 & resid\_post & 1e-5  & 98.6 & 1110.9  & 40.46   & 0.993 & 0.988 & 6.762 & 6.763 & 6.908 & 99.23  & 100    & \href{https://huggingface.co/prisma-multimodal/sparse-autoencoder-clip-b-32-sae-vanilla-x64-layer-0-hook_resid_post-l1-1e-05}{link} \\
1 & mlp\_out    & 1e-5  & 98.4 & 1476.8  & 97.82   & 0.992 & 0.994 & 6.762 & 6.762 & 6.889 & 99.40  & 100    & \href{https://huggingface.co/prisma-multimodal/sparse-autoencoder-clip-b-32-sae-vanilla-x64-layer-1-hook_mlp_out-l1-1e-05}{link} \\
1 & resid\_post & 1e-5  & 98.3 & 1508.4  & 27.39   & 0.991 & 0.989 & 6.762 & 6.763 & 6.908 & 99.02  & 100    & \href{https://huggingface.co/prisma-multimodal/sparse-autoencoder-clip-b-32-sae-vanilla-x64-layer-1-hook_resid_post-l1-1e-05}{link} \\
2 & mlp\_out    & 1e-5  & 98.0 & 1799.7  & 376.0   & 0.992 & 0.998 & 6.762 & 6.762 & 6.803 & 99.44  & 100    & \href{https://huggingface.co/prisma-multimodal/sparse-autoencoder-clip-b-32-sae-vanilla-x64-layer-2-hook_mlp_out-l1-1e-05}{link} \\
2 & resid\_post & 5e-5  & 90.6 & 717.84  & 10.11   & 0.944 & 0.960 & 6.762 & 6.767 & 6.908 & 96.34  & 100    & \href{https://huggingface.co/prisma-multimodal/sparse-autoencoder-clip-b-32-sae-vanilla-x64-layer-2-hook_resid_post-l1-5e-05}{link} \\
3 & mlp\_out    & 1e-5  & 98.1 & 1893.4  & 648.2   & 0.992 & 0.999 & 6.762 & 6.762 & 6.784 & 99.54  & 100    & \href{https://huggingface.co/prisma-multimodal/sparse-autoencoder-clip-b-32-sae-vanilla-x64-layer-3-hook_mlp_out-l1-1e-05}{link} \\
3 & resid\_post & 1e-5  & 98.1 & 2053.9  & 77.90   & 0.989 & 0.996 & 6.762 & 6.762 & 6.908 & 99.79  & 100    & \href{https://huggingface.co/prisma-multimodal/sparse-autoencoder-clip-b-32-sae-vanilla-x64-layer-3-hook_resid_post-l1-1e-05}{link} \\
4 & mlp\_out    & 1e-5  & 98.1 & 1901.2  & 1115.0  & 0.993 & 0.999 & 6.762 & 6.762 & 6.786 & 99.55  & 100    & \href{https://huggingface.co/prisma-multimodal/sparse-autoencoder-clip-b-32-sae-vanilla-x64-layer-4-hook_mlp_out-l1-1e-05}{link} \\
4 & resid\_post & 1e-5  & 98.0 & 2068.3  & 156.7   & 0.989 & 0.997 & 6.762 & 6.762 & 6.908 & 99.74  & 100    & \href{https://huggingface.co/prisma-multimodal/sparse-autoencoder-clip-b-32-sae-vanilla-x64-layer-4-hook_resid_post-l1-1e-05}{link} \\
5 & mlp\_out    & 1e-5  & 98.2 & 1761.5  & 1259.0  & 0.993 & 0.999 & 6.762 & 6.762 & 6.797 & 99.76  & 100    & \href{https://huggingface.co/prisma-multimodal/sparse-autoencoder-clip-b-32-sae-vanilla-x64-layer-5-hook_mlp_out-l1-1e-05}{link} \\
5 & resid\_post & 1e-5  & 98.1 & 1953.8  & 228.5   & 0.990 & 0.997 & 6.762 & 6.762 & 6.908 & 99.80  & 100    & \href{https://huggingface.co/prisma-multimodal/sparse-autoencoder-clip-b-32-sae-vanilla-x64-layer-5-hook_resid_post-l1-1e-05}{link} \\
6 & mlp\_out    & 1e-5  & 98.3 & 1598.0  & 1337.0  & 0.993 & 0.999 & 6.762 & 6.762 & 6.789 & 99.83  & 100    & \href{https://huggingface.co/prisma-multimodal/sparse-autoencoder-clip-b-32-sae-vanilla-x64-layer-6-hook_mlp_out-l1-1e-05}{link} \\
6 & resid\_post & 1e-5  & 98.2 & 1717.5  & 321.3   & 0.991 & 0.996 & 6.762 & 6.762 & 6.908 & 99.93  & 100    & \href{https://huggingface.co/prisma-multimodal/sparse-autoencoder-clip-b-32-sae-vanilla-x64-layer-6-hook_resid_post-l1-1e-05}{link} \\
7 & mlp\_out    & 1e-5  & 98.2 & 1535.3  & 1300.0  & 0.992 & 0.999 & 6.762 & 6.762 & 6.796 & 100.17 & 100    & \href{https://huggingface.co/prisma-multimodal/sparse-autoencoder-clip-b-32-sae-vanilla-x64-layer-7-hook_mlp_out-l1-1e-05}{link} \\
7 & resid\_post & 1e-5  & 98.2 & 1688.4  & 494.3   & 0.991 & 0.995 & 6.762 & 6.761 & 6.908 & 100.24 & 100    & \href{https://huggingface.co/prisma-multimodal/sparse-autoencoder-clip-b-32-sae-vanilla-x64-layer-7-hook_resid_post-l1-1e-05}{link} \\
8 & mlp\_out    & 1e-5  & 97.8 & 1074.5  & 1167.0  & 0.990 & 0.998 & 6.762 & 6.761 & 6.793 & 100.57 & 100    & \href{https://huggingface.co/prisma-multimodal/sparse-autoencoder-clip-b-32-sae-vanilla-x64-layer-8-hook_mlp_out-l1-1e-05}{link} \\
8 & resid\_post & 1e-5  & 98.2 & 1570.8  & 791.3   & 0.991 & 0.992 & 6.762 & 6.761 & 6.908 & 100.41 & 100    & \href{https://huggingface.co/prisma-multimodal/sparse-autoencoder-clip-b-32-sae-vanilla-x64-layer-8-hook_resid_post-l1-1e-05}{link} \\
9 & mlp\_out    & 1e-5  & 97.6 & 856.68  & 1076.0  & 0.989 & 0.998 & 6.762 & 6.762 & 6.792 & 100.28 & 100    & \href{https://huggingface.co/prisma-multimodal/sparse-autoencoder-clip-b-32-sae-vanilla-x64-layer-9-hook_mlp_out-l1-1e-05}{link} \\
9 & resid\_post & 1e-5  & 98.2 & 1533.5  & 1053.0  & 0.991 & 0.989 & 6.762 & 6.761 & 6.908 & 100.32 & 100    & \href{https://huggingface.co/prisma-multimodal/sparse-autoencoder-clip-b-32-sae-vanilla-x64-layer-9-hook_resid_post-l1-1e-05}{link} \\
10 & mlp\_out   & 1e-5  & 98.1 & 788.49  & 965.5   & 0.991 & 0.998 & 6.762 & 6.762 & 6.772 & 101.50 & 99.80  & \href{https://huggingface.co/prisma-multimodal/sparse-autoencoder-clip-b-32-sae-vanilla-x64-layer-10-hook_mlp_out-l1-1e-05}{link} \\
10 & resid\_post& 1e-5  & 98.4 & 1292.6  & 1010.0  & 0.992 & 0.987 & 6.762 & 6.760 & 6.908 & 100.83 & 99.99  & \href{https://huggingface.co/prisma-multimodal/sparse-autoencoder-clip-b-32-sae-vanilla-x64-layer-10-hook_resid_post-l1-1e-05}{link} \\
11 & mlp\_out   & 5e-5  & 89.7 & 748.14  & 745.5   & 0.972 & 0.993 & 6.762 & 6.759 & 6.768 & 135.77 & 100    & \href{https://huggingface.co/prisma-multimodal/sparse-autoencoder-clip-b-32-sae-vanilla-x64-layer-11-hook_mlp_out-l1-5e-05}{link} \\
11 & resid\_post& 1e-5  & 98.4 & 1405.0  & 1189.0  & 0.993 & 0.987 & 6.762 & 6.765 & 6.908 & 98.03  & 99.99  & \href{https://huggingface.co/prisma-multimodal/sparse-autoencoder-clip-b-32-sae-vanilla-x64-layer-11-hook_resid_post-l1-1e-05}{link} \\
\hline
\end{tabular}
}

\subsubsection*{Vanilla SAEs (CLS only)}
\captionof{table}{CLIP-ViT-B-32 vanilla sparse autoencoder performance metrics for CLS tokens.}
\label{table:cls_only}
\resizebox{\textwidth}{!}{%
\begin{tabular}{|c|c|c|c|c|c|c|c|c|c|c|c|l|}
\hline
\textbf{Layer} & \textbf{Sublayer} & \textbf{l1 coeff.} & \textbf{\% Explained var.} & \textbf{Avg CLS L0} & \textbf{Cos sim} & \textbf{Recon cos sim} & \textbf{CE} & \textbf{Recon CE} & \textbf{Zero abl CE} & \textbf{\% CE recovered} & \textbf{\% Alive features} & \textbf{Model} \\
\hline
0  & resid\_post & 2e-8   & 82  & 934.83 & 0.98008 & 0.99995 & 6.7622 & 6.7622 & 6.9084 & 99.9984 & 4.33   & \href{https://huggingface.co/Prisma-Multimodal/imagenet-sweep-vanilla-x64-CLS_0-hook_resid_post-936.799987792969-82}{link} \\
1  & resid\_post & 8e-6   & 85  & 314.13 & 0.97211 & 0.99994 & 6.7622 & 6.7622 & 6.9083 & 100.00   & 2.82   & \href{https://huggingface.co/Prisma-Multimodal/imagenet-sweep-vanilla-x64-CLS_1-hook_resid_post-314.175018310547-85}{link} \\
2  & resid\_post & 9e-8   & 96  & 711.84 & 0.98831 & 0.99997 & 6.7622 & 6.7622 & 6.9083 & 99.9977  & 2.54   & \href{https://huggingface.co/Prisma-Multimodal/imagenet-sweep-vanilla-x64-CLS_2-hook_resid_post-711.121887207031-96}{link} \\
3  & resid\_post & 1e-8   & 95  & 687.41 & 0.98397 & 0.99994 & 6.7622 & 6.7622 & 6.9085 & 99.9998  & 4.49   & \href{https://huggingface.co/Prisma-Multimodal/imagenet-sweep-vanilla-x64-CLS_3-hook_resid_post-686.334411621094-95}{link} \\
4  & resid\_post & 9e-8   & 95  & 681.08 & 0.98092 & 0.99988 & 6.7622 & 6.7622 & 6.9082 & 100.00   & 15.75  & \href{https://huggingface.co/Prisma-Multimodal/imagenet-sweep-vanilla-x64-CLS_4-hook_resid_post-682.543762207031-95}{link} \\
5  & resid\_post & 1e-7   & 94  & 506.77 & 0.97404 & 0.99966 & 6.7622 & 6.7622 & 6.9081 & 99.9911  & 16.80  & \href{https://huggingface.co/Prisma-Multimodal/imagenet-sweep-vanilla-x64-CLS_5-hook_resid_post-510.356262207031-94}{link} \\
6  & resid\_post & 1e-8   & 92  & 423.70 & 0.96474 & 0.99913 & 6.7622 & 6.7622 & 6.9083 & 99.9971  & 29.46  & \href{https://huggingface.co/Prisma-Multimodal/imagenet-sweep-vanilla-x64-CLS_6-hook_resid_post-430.556243896484-92}{link} \\
7  & resid\_post & 2e-6   & 88  & 492.68 & 0.93899 & 0.99737 & 6.7622 & 6.7622 & 6.9082 & 99.9583  & 51.68  & \href{https://huggingface.co/Prisma-Multimodal/imagenet-sweep-vanilla-x64-CLS_7-hook_resid_post-492.959381103516-88}{link} \\
8  & resid\_post & 4e-8   & 76  & 623.01 & 0.89168 & 0.99110 & 6.7622 & 6.7625 & 6.9087 & 99.7631  & 82.07  & \href{https://huggingface.co/Prisma-Multimodal/imagenet-sweep-vanilla-x64-CLS_8-hook_resid_post-635.018737792969-76}{link} \\
9  & resid\_post & 1e-12  & 74  & 521.90 & 0.87076 & 0.98191 & 6.7622 & 6.7628 & 6.9083 & 99.5425  & 93.68  & \href{https://huggingface.co/Prisma-Multimodal/imagenet-sweep-vanilla-x64-CLS_9-hook_resid_post-518.621887207031-74}{link} \\
10 & resid\_post & 3e-7   & 74  & 533.94 & 0.87646 & 0.96514 & 6.7622 & 6.7635 & 6.9082 & 99.1070  & 99.98  & \href{https://huggingface.co/Prisma-Multimodal/imagenet-sweep-vanilla-x64-CLS_10-hook_resid_post-552.512512207031-74}{link} \\
11 & resid\_post & 1e-8   & 65  & 386.09 & 0.81890 & 0.89607 & 6.7622 & 6.7853 & 6.9086 & 84.1918  & 99.996 & \href{https://huggingface.co/Prisma-Multimodal/imagenet-sweep-vanilla-x64-CLS_11-hook_resid_post-383.75-65}{link} \\
\hline
\end{tabular}
}

\subsubsection*{Top K SAEs (CLS only, $k=64$)}
\captionof{table}{CLIP-ViT-B-32 top-$k$ sparse autoencoder performance metrics for CLS tokens ($k=64$).}
\label{table:top_k_cls_64}
\footnotesize
\resizebox{\textwidth}{!}{%
\begin{tabular}{|c|c|c|c|c|c|c|c|c|c|c|c|l|}
\hline
\textbf{Layer} & \textbf{Sublayer} & \textbf{\% Explained var.} & \textbf{Avg CLS L0} & \textbf{Cos sim} & \textbf{Recon cos sim} & \textbf{CE} & \textbf{Recon CE} & \textbf{Zero abl CE} & \textbf{\% CE recovered} & \textbf{\% Alive features} & \textbf{Model} \\
\hline
0  & resid\_post & 90 & 64 & 0.98764 & 0.99998 & 6.7622 & 6.7622 & 6.9084 & 99.995 & 46.80 & \href{https://huggingface.co/Prisma-Multimodal/sae-top_k-64-cls_only-layer_0-hook_resid_post}{link} \\
1  & resid\_post & 96 & 64 & 0.99429 & 0.99999 & 6.7622 & 6.7622 & 6.9083 & 100.00 & 4.86  & \href{https://huggingface.co/Prisma-Multimodal/sae-top_k-64-cls_only-layer_1-hook_resid_post}{link} \\
2  & resid\_post & 96 & 64 & 0.99000 & 0.99998 & 6.7622 & 6.7622 & 6.9083 & 100.00 & 5.50  & \href{https://huggingface.co/Prisma-Multimodal/sae-top_k-64-cls_only-layer_2-hook_resid_post}{link} \\
3  & resid\_post & 95 & 64 & 0.98403 & 0.99995 & 6.7622 & 6.7622 & 6.9085 & 100.00 & 5.21  & \href{https://huggingface.co/Prisma-Multimodal/sae-top_k-64-cls_only-layer_3-hook_resid_post}{link} \\
4  & resid\_post & 94 & 64 & 0.97485 & 0.99986 & 6.7621 & 6.7622 & 6.9082 & 99.998 & 6.81  & \href{https://huggingface.co/Prisma-Multimodal/sae-top_k-64-cls_only-layer_4-hook_resid_post}{link} \\
5  & resid\_post & 93 & 64 & 0.96985 & 0.99962 & 6.7622 & 6.7622 & 6.9081 & 99.997 & 21.89 & \href{https://huggingface.co/Prisma-Multimodal/sae-top_k-64-cls_only-layer_5-hook_resid_post}{link} \\
6  & resid\_post & 92 & 64 & 0.96401 & 0.99912 & 6.7622 & 6.7622 & 6.9083 & 100.00 & 28.81 & \href{https://huggingface.co/Prisma-Multimodal/sae-top_k-64-cls_only-layer_6-hook_resid_post}{link} \\
7  & resid\_post & 90 & 64 & 0.95057 & 0.99797 & 6.7622 & 6.7621 & 6.9082 & 100.03 & 65.84 & \href{https://huggingface.co/Prisma-Multimodal/sae-top_k-64-cls_only-layer_7-hook_resid_post}{link} \\
8  & resid\_post & 87 & 64 & 0.93029 & 0.99475 & 6.7622 & 6.7620 & 6.9087 & 100.11 & 93.75 & \href{https://huggingface.co/Prisma-Multimodal/sae-top_k-64-cls_only-layer_8-hook_resid_post}{link} \\
9  & resid\_post & 85 & 64 & 0.91814 & 0.98865 & 6.7622 & 6.7616 & 6.9083 & 100.43 & 98.90 & \href{https://huggingface.co/Prisma-Multimodal/sae-top_k-64-cls_only-layer_9-hook_resid_post}{link} \\
10 & resid\_post & 86 & 64 & 0.93072 & 0.97929 & 6.7622 & 6.7604 & 6.9082 & 101.19 & 94.55 & \href{https://huggingface.co/Prisma-Multimodal/sae-top_k-64-cls_only-layer_10-hook_resid_post}{link} \\
11 & resid\_post & 84 & 64 & 0.91880 & 0.94856 & 6.7622 & 6.7578 & 6.9086 & 102.97 & 97.99 & \href{https://huggingface.co/Prisma-Multimodal/sae-top_k-64-cls_only-layer_11-hook_resid_post}{link} \\
\hline
\end{tabular}
}

\subsubsection*{Vanilla SAEs (Spatial Patches)}
\label{sec:spatial_patches}
\captionof{table}{CLIP-ViT-B-32 vanilla sparse autoencoder performance metrics for spatial patches.}
\footnotesize
\resizebox{\textwidth}{!}{%
\begin{tabular}{|c|c|c|c|c|c|c|c|c|c|c|c|l|}
\hline
\textbf{Layer} & \textbf{Sublayer} & \textbf{l1 coeff.} & \textbf{\% Explained var.} & \textbf{Avg L0} & \textbf{Cos sim} & \textbf{Recon cos sim} & \textbf{CE} & \textbf{Recon CE} & \textbf{Zero abl CE} & \textbf{\% CE recovered} & \textbf{\% Alive features} & \textbf{Model} \\
\hline
0  & resid\_post & 1e-12 & 99 & 989.19  & 0.99 & 0.99 & 6.7621 & 6.7621 & 6.9084 & 99.9981 & 100.00 & \href{https://huggingface.co/Prisma-Multimodal/imagenet-sweep-vanilla-x64-Spatial_max_0-hook_resid_post-989.203430175781-99}{link} \\
1  & resid\_post & 3e-11 & 99 & 757.83  & 0.99 & 0.99 & 6.7622 & 6.7622 & 6.9083 & 99.9969 & 45.39  & \href{https://huggingface.co/Prisma-Multimodal/imagenet-sweep-vanilla-x64-Spatial_max_1-hook_resid_post-757.82958984375-99}{link} \\
2  & resid\_post & 4e-12 & 99 & 1007.89 & 0.99 & 0.99 & 6.7622 & 6.7622 & 6.9083 & 100.00  & 97.93  & \href{https://huggingface.co/Prisma-Multimodal/imagenet-sweep-vanilla-x64-Spatial_max_2-hook_resid_post-1007.89801025391-99}{link} \\
3  & resid\_post & 2e-8  & 99 & 935.06  & 0.99 & 0.99 & 6.7622 & 6.7622 & 6.9085 & 99.9882 & 100.00 & \href{https://huggingface.co/Prisma-Multimodal/imagenet-sweep-vanilla-x64-Spatial_max_3-hook_resid_post-935.601989746094-99}{link} \\
4  & resid\_post & 3e-8  & 99 & 965.15  & 0.99 & 0.99 & 6.7622 & 6.7622 & 6.9082 & 99.9842 & 100.00 & \href{https://huggingface.co/Prisma-Multimodal/imagenet-sweep-vanilla-x64-Spatial_max_4-hook_resid_post-965.410095214844-99}{link} \\
5  & resid\_post & 1e-8  & 99 & 966.38  & 0.99 & 0.99 & 6.7622 & 6.7622 & 6.9081 & 99.9961 & 100.00 & \href{https://huggingface.co/Prisma-Multimodal/imagenet-sweep-vanilla-x64-Spatial_max_5-hook_resid_post-964.674072265625-99}{link} \\
6  & resid\_post & 1e-8  & 99 & 1006.62 & 0.99 & 0.99 & 6.7622 & 6.7622 & 6.9083 & 100.00  & 99.97  & \href{https://huggingface.co/Prisma-Multimodal/imagenet-sweep-vanilla-x64-Spatial_max_6-hook_resid_post-1006.57165527344-99}{link} \\
7  & resid\_post & 1e-8  & 99 & 984.19  & 0.99 & 0.99 & 6.7622 & 6.7622 & 6.9082 & 100.00  & 100.00 & \href{https://huggingface.co/Prisma-Multimodal/imagenet-sweep-vanilla-x64-Spatial_max_7-hook_resid_post-984.1376953125-99}{link} \\
8  & resid\_post & 3e-8  & 99 & 965.12  & 0.99 & 1.00 & 6.7622 & 6.7622 & 6.9087 & 100.00  & 92.37  & \href{https://huggingface.co/Prisma-Multimodal/imagenet-sweep-vanilla-x64-Spatial_max_8-hook_resid_post-965.125-99}{link} \\
9  & resid\_post & 9e-8  & 99 & 854.92  & 0.99 & 1.00 & 6.7622 & 6.7622 & 6.9083 & 99.9991 & 85.43  & \href{https://huggingface.co/Prisma-Multimodal/imagenet-sweep-vanilla-x64-Spatial_max_9-hook_resid_post-854.891540527344-99}{link} \\
10 & resid\_post & 1e-4  & 72 & 88.80  & 0.84 & 0.97 & 6.7621 & 6.7638 & 6.9082 & 98.85  & 100.00  & \href{https://huggingface.co/Prisma-Multimodal/imagenet-sweep-vanilla-x64-Spatial_max_10-hook_resid_post-90.03951-72}{link} \\
11 & resid\_post & 3e-7  & 99 & 829.09  & 0.99 & 1.00 & 6.7622 & 6.7622 & 6.9086 & 100.00  & 55.71  & \href{https://huggingface.co/Prisma-Multimodal/imagenet-sweep-vanilla-x64-Spatial_max_11-hook_resid_post-829.0498046875-99}{link} \\
\hline
\end{tabular}
}

\subsubsection*{Top K Transcoders (All Patches)}
\captionof{table}{CLIP Top-K transcoder performance metrics for all patches.}
\label{tab:clip_transcoders}
\footnotesize
\resizebox{\textwidth}{!}{%
\begin{tabular}{|c|c|c|c|c|c|c|c|c|c|l|}
\hline
\textbf{Layer} & \textbf{Block} & \textbf{\% Explained var.} & \textbf{k} & \textbf{Avg CLS L0} & \textbf{Cos sim} & \textbf{CE} & \textbf{Recon CE} & \textbf{Zero abl CE} & \textbf{\% CE recovered} & \textbf{Model} \\
\hline
0 & MLP & 96 & 768 & 767 & 0.9655 & 6.7621 & 6.7684 & 6.8804 & 94.68 & \href{https://huggingface.co/Prisma-Multimodal/CLIP-transcoder-topk-768-x64-all_patches_0-mlp-96}{link} \\
1 & MLP & 94 & 256 & 255 & 0.9406 & 6.7621 & 6.7767 & 6.8816 & 87.78 & \href{https://huggingface.co/Prisma-Multimodal/CLIP-transcoder-topk-256-x64-all_patches_1-mlp-94}{link} \\
2 & MLP & 93 & 1024 & 475 & 0.9758 & 6.7621 & 6.7681 & 6.7993 & 83.92 & \href{https://huggingface.co/Prisma-Multimodal/CLIP-transcoder-topk-1024-x64-all_patches_2-mlp-93}{link} \\
3 & MLP & 90 & 1024 & 825 & 0.9805 & 6.7621 & 6.7642 & 6.7999 & 94.42 & \href{https://huggingface.co/Prisma-Multimodal/CLIP-transcoder-topk-1024-x64-all_patches_3-mlp-90}{link} \\
4 & MLP & 76 & 512 & 29  & 0.9830 & 6.7621 & 6.7636 & 6.8080 & 96.76 & \href{https://huggingface.co/Prisma-Multimodal/CLIP-transcoder-topk-512-x64-all_patches_4-mlp-76}{link} \\
5 & MLP & 91 & 1024 & 1017& 0.9784 & 6.7621 & 6.7643 & 6.8296 & 96.82 & \href{https://huggingface.co/Prisma-Multimodal/CLIP-transcoder-topk-1024-x64-all_patches_5-mlp-91}{link} \\
6 & MLP & 94 & 1024 & 924 & 0.9756 & 6.7621 & 6.7630 & 6.8201 & 98.40 & \href{https://huggingface.co/Prisma-Multimodal/CLIP-transcoder-topk-1024-x64-all_patches_6-mlp-94}{link} \\
7 & MLP & 97 & 1024 & 1010& 0.9629 & 6.7621 & 6.7631 & 6.8056 & 97.68 & \href{https://huggingface.co/Prisma-Multimodal/CLIP-transcoder-topk-1024-x64-all_patches_7-mlp-97}{link} \\
8 & MLP & 98 & 1024 & 1023& 0.9460 & 6.7621 & 6.7630 & 6.8017 & 97.70 & \href{https://huggingface.co/Prisma-Multimodal/CLIP-transcoder-topk-1024-x64-all_patches_8-mlp-98}{link} \\
9 & MLP & 98 & 1024 & 1023& 0.9221 & 6.7621 & 6.7630 & 6.7875 & 96.50 & \href{https://huggingface.co/Prisma-Multimodal/CLIP-transcoder-topk-1024-x64-all_patches_9-mlp-98}{link} \\
10 & MLP & 97& 1024 & 1019& 0.9334 & 6.7621 & 6.7636 & 6.7860 & 93.95 & \href{https://huggingface.co/Prisma-Multimodal/CLIP-transcoder-topk-1024-x64-all_patches_10-mlp-97}{link} \\
\hline
\end{tabular}
}

\subsection{DINO-B SAEs}


\subsubsection*{Vanilla (All patches)}
\captionof{table}{DINO Vanilla sparse autoencoder performance metrics for all patches (CLS + spatial).}
\label{table:dino_vanilla}
\footnotesize
\resizebox{\textwidth}{!}{%
\begin{tabular}{|c|c|c|c|c|c|c|c|c|c|l|}
\hline
\textbf{Layer} & \textbf{Sublayer} & \textbf{Avg L0.} & \textbf{\% Explained var.} & \textbf{Avg CLS L0} & \textbf{Cos sim} & \textbf{CE} & \textbf{Recon CE} & \textbf{Zero abl CE} & \textbf{\% CE Recovered} & \textbf{Model} \\
\hline
0  & resid\_post & 507 & 98 & 347 & 0.95009 & 1.885033 & 1.936518 & 7.2714 & 99.04 & \href{https://huggingface.co/Prisma-Multimodal/DINO-vanilla-x64-all_patches_0-resid_post-507-98}{link} \\
1  & resid\_post & 549 & 95 & 959 & 0.93071 & 1.885100 & 1.998274 & 7.2154 & 97.88 & \href{https://huggingface.co/Prisma-Multimodal/DINO-vanilla-x64-all_patches_1-resid_post-549-95}{link} \\
2  & resid\_post & 812 & 95 & 696 & 0.9560 & 1.885134 & 2.006115 & 7.201461 & 97.72 & \href{https://huggingface.co/Prisma-Multimodal/DINO-vanilla-x64-all_patches_2-resid_post-661-95}{link} \\
3  & resid\_post & 989 & 95 & 616 & 0.96315 & 1.885131 & 1.961913 & 7.2068 & 98.56 & \href{https://huggingface.co/Prisma-Multimodal/DINO-vanilla-x64-all_patches_3-resid_post-989-95}{link} \\
4  & resid\_post & 876 & 99 & 845 & 0.99856 & 1.885224 & 1.883169 & 7.1636 & 100.04 & \href{https://huggingface.co/Prisma-Multimodal/DINO-vanilla-x64-all_patches_4-resid_post-876-99}{link} \\
5  & resid\_post & 1001 & 98 & 889 & 0.99129 & 1.885353 & 1.875520 & 7.1412 & 100.19 & \href{https://huggingface.co/Prisma-Multimodal/DINO-vanilla-x64-all_patches_5-resid_post-1001-98}{link} \\
6  & resid\_post & 962 & 99 & 950 & 0.99945 & 1.885239 & 1.872594 & 7.1480 & 100.24 & \href{https://huggingface.co/Prisma-Multimodal/DINO-vanilla-x64-all_patches_6-resid_post-962-99}{link} \\
7  & resid\_post & 1086 & 98 & 1041 & 0.99341 & 1.885371 & 1.869443 & 7.1694 & 100.30 & \href{https://huggingface.co/Prisma-Multimodal/DINO-vanilla-x64-all_patches_7-resid_post-1086-98}{link} \\
8  & resid\_post & 530 & 90 & 529 & 0.9475 & 1.885511 & 1.978638 & 7.1315 & 98.22 & \href{https://huggingface.co/Prisma-Multimodal/DINO-vanilla-x64-all_patches_8-resid_post-530-90}{link} \\
9  & resid\_post & 1105 & 99 & 1090 & 0.99541 & 1.885341 & 1.894026 & 7.0781 & 99.83 & \href{https://huggingface.co/Prisma-Multimodal/DINO-vanilla-x64-all_patches_9-resid_post-1105-99}{link} \\
10  & resid\_post & 835 & 99 & 839 & 0.99987 & 1.885371 & 1.884487 & 7.3606 & 100.02 & \href{https://huggingface.co/Prisma-Multimodal/DINO-vanilla-x64-all_patches_10-resid_post-835-99}{link} \\
11 & resid\_post & 1085 & 99 & 1084 & 0.99673 & 1.885370 & 1.911608 & 6.9078 & 99.48 & \href{https://huggingface.co/Prisma-Multimodal/DINO-vanilla-x64-all_patches_11-resid_post-1085-99}{link} \\
\hline
\end{tabular}
}

\end{document}